\PassOptionsToPackage{table,xcdraw}{xcolor}
\documentclass[titlefont=sans,logoposition=abstract]{team_template}

\setteamabstractlogoautosize{false}
\setteamabstractlogoscale{1.0}
\setteamlogofile{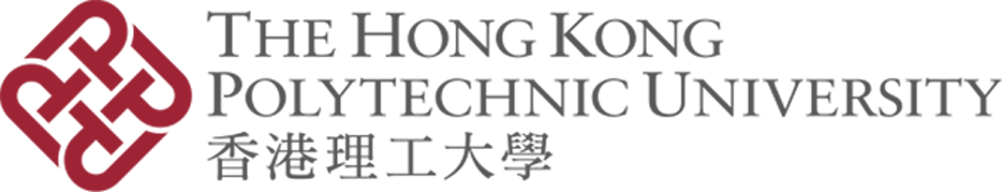}
\setteamtopseparatorcolor{teamred}

\usepackage{amsmath,amssymb}
\usepackage{tabularx}
\usepackage{float}
\usepackage{algorithm}
\usepackage{algorithmic}

\ifPDFTeX
\else
  \AtBeginDvi{}
\fi

\newcommand{\srval}[2]{\mbox{#1{\scriptsize$\pm$#2}}}
\newcommand{\bval}[2]{\textbf{\boldmath\srval{#1}{#2}}}
\newcommand{\uval}[2]{\underline{\srval{#1}{#2}}}

\hypersetup{
  pdftitle={PhysVAM: Physics-oriented Future Value Action Model for Dynamic Object Manipulation},
  pdfauthor={Yuxiang Yang et al.}
}

\title{PhysVAM: Physics-oriented Future Value Action Model for Dynamic Object Manipulation}
\author[1]{Yuxiang Yang}
\author[2]{Jingbo Zhang}
\author[1]{Jiaru Zhong}
\author[1]{Junliang Chen}
\author[1]{Yirui Zhou}
\author[3]{Guangming Wang}
\author[2]{Haitao Lin}
\author[1]{Huaiyuan Xu}
\author[1]{Yi Wang}
\author[1]{Lap-Pui Chau}
\affiliation[1]{Hong Kong Polytechnic University}
\affiliation[2]{Robotics X, Tencent}
\affiliation[3]{University of Cambridge}
\abstract{Predicting future scene or object states does not necessarily reveal how robot--object interactions will advance or hinder a task. This challenge is particularly pronounced in dynamic robotic manipulation, where objects continue moving during perception and execution. Effective action generation therefore requires not only modeling object dynamics but also anticipating how future interactions will shape task progress. To this end, we propose \textbf{PhysVAM}, a Physics-oriented Future Value Action Model that represents anticipated interactions through their effects on future task progress. This value-based formulation provides task-directed foresight without reconstructing future visual states. Conditioned on current observations and historical object and robot motion, PhysVAM jointly learns action generation and multi-horizon future-value prediction. The value branch learns horizon-specific latent representations through future task-value prediction from successful and failed trajectories. These future-value representations are integrated into the action policy through gated cross-attention, guiding action generation toward interactions that facilitate task success. Extensive simulation and real-world experiments demonstrate superior task success, with absolute success-rate improvements of 27.0 and 27.5 percentage points over the strongest baselines in simulation and real-world tasks, respectively.
}
\date{November 19, 2026}
\metadata[Project Page]{\url{https://metarobo.github.io/physvam.github.io/}}

\begin{document}
\maketitle
\begin{figure}[!t]
    \centering
    \includegraphics[width=\linewidth]{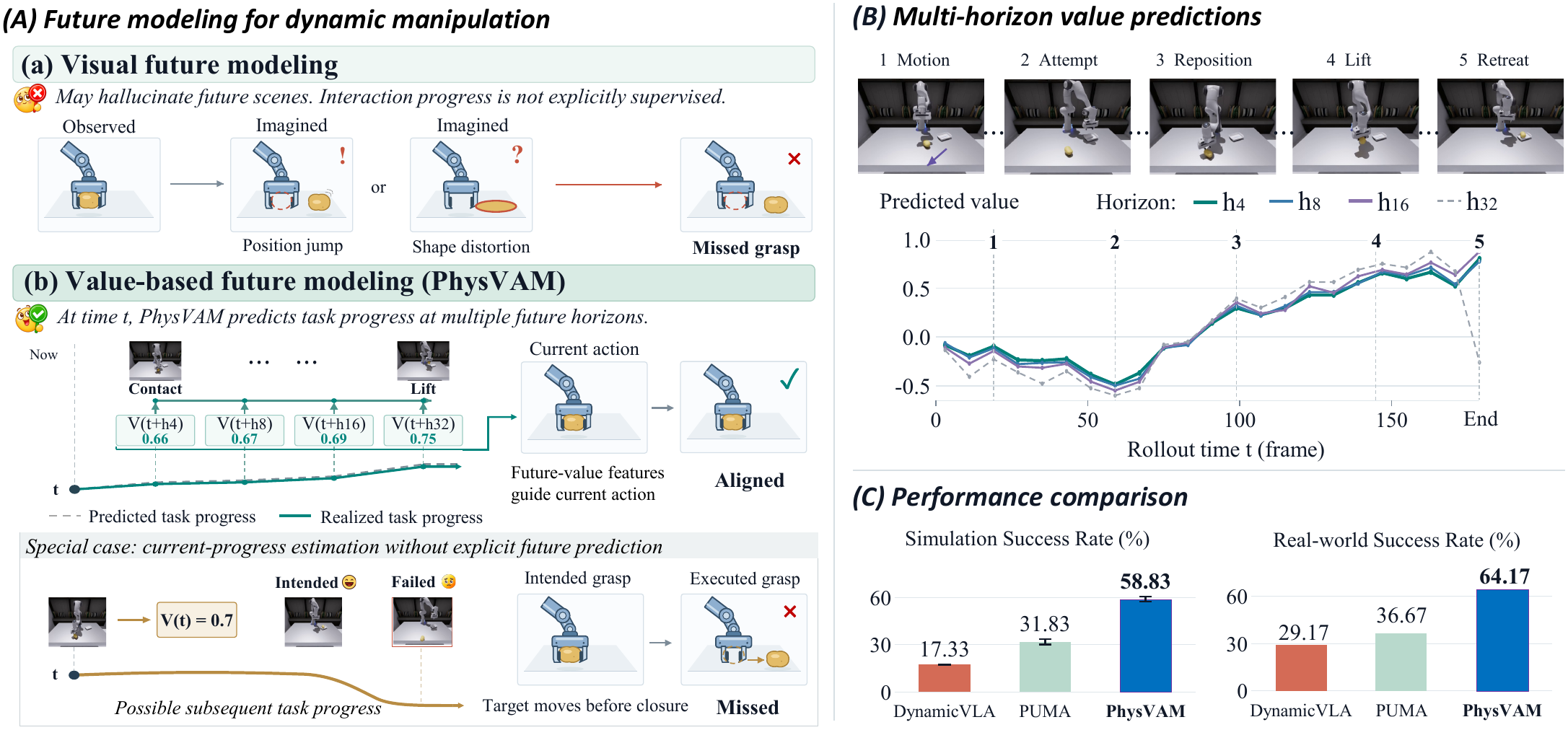}
    \caption{
    \textbf{Overview of PhysVAM.}
    \textbf{(A)} Comparison of visual and value-based future modeling.  Visual prediction may suffer from geometric inconsistencies and lacks explicit task-progress supervision.
    In contrast, PhysVAM predicts multi-horizon future values
    to guide action generation.  The $h=0$ special case illustrates how favorable current-progress estimates may still precede unsuccessful interactions.
    \textbf{(B)} Multi-horizon value predictions decrease following
    a missed grasp and increase during subsequent recovery
    and successful lifting, reflecting changes in anticipated task progress.
    \textbf{(C)} PhysVAM achieves the highest success rates
    among the compared methods in both simulation
    and real-world tasks.}
    \label{fig:physvam_motivation}
\end{figure}

\section{Introduction}
\label{sec:introduction}

Robotic manipulation is increasingly moving beyond static tabletop settings toward dynamic environments, where objects continue to move during robot perception and action execution~\cite{brohan2023rt,xie2026dynamicvla}. Successful manipulation therefore requires actions to remain coordinated with evolving robot--object interactions. Changes in object motion and contact timing can alter whether an intended interaction will succeed before its execution is complete. A central challenge is thus to learn predictive representations that anticipate how future interactions may affect task progress and guide timely action generation.

Recent approaches pursue visual future modeling through future video generation~\cite{yuan2026fast,lou2026dynamicwam} and visual-feature prediction~\cite{fang2026towards,sun2026vla}. These methods provide predictive context by modeling how scenes and objects may evolve. However, as illustrated in Fig.~\ref{fig:physvam_motivation}(A-a), generated futures may exhibit geometric distortions and inconsistent contact predictions, compromising their reliability for physical interaction~\cite{jiang2026robowm}. More fundamentally, predicting future visual or object states alone does not explicitly characterize how anticipated robot--object interactions will advance or hinder task progress. These limitations motivate a future modeling objective that directly reflects the task consequences of physical interactions.

We approach future modeling from a task-value perspective, representing anticipated robot--object interactions through their effects on future task progress. Rather than reconstructing future visual states, this perspective models how task progress is expected to evolve. Existing value-based approaches have demonstrated the usefulness of task evaluation for policy improvement~\cite{lv2026viva,intelligence2025pi,fei2026wcm}. However, current-value assessment does not explicitly characterize how task progress may evolve through subsequent interactions. For instance, a favorable current value may still be followed by a missed grasp as the target moves before gripper closure. Predicting task values across multiple future horizons instead provides temporally structured supervision for learning task-directed representations that guide action generation, as illustrated in Fig.~\ref{fig:physvam_motivation}(A-b).

To instantiate this perspective, we propose \textbf{PhysVAM}, a \underline{\textbf{Phys}}ics-oriented Future \underline{\textbf{V}}alue \underline{\textbf{A}}ction \underline{\textbf{M}}odel for dynamic object manipulation. PhysVAM leverages current observations and historical object and robot motion to model evolving interaction context, while jointly learning action generation and multi-horizon future-value prediction. The value branch predicts task values at multiple future horizons and learns corresponding latent representations from both successful and failed trajectories. These future-value representations capture anticipated task progress associated with robot--object interactions and are integrated into the action policy through gated cross-attention, guiding action generation toward interactions that promote task success. As illustrated in Fig.~\ref{fig:physvam_motivation}(B), the predicted future values can reflect changes in task progress as interactions unfold.

Our main contributions are summarized as follows:
\begin{itemize}
    \item We present a task-value perspective on future modeling for dynamic object manipulation, leveraging multi-horizon future-value prediction to learn task-directed representations for action guidance.

    \item We propose PhysVAM, a physics-oriented framework that integrates historical object and robot motion understanding with multi-horizon future-value guidance for action generation.

    \item Extensive simulation and real-world experiments demonstrate improved task success and execution efficiency, with absolute success-rate gains of 27.0 and 27.5 percentage points over the strongest baselines, respectively. Integrating our value branch into other action models yields further improvements, demonstrating the transferability of future-value guidance.
\end{itemize}

\section{Related Work}
\label{sec:related_work}

\noindent\textbf{Vision--Language--Action and World--Action Models.}~
Vision-language-action (VLA) models have advanced generalist robot manipulation by learning multimodal policies that map visual observations and language instructions to robot actions ~\cite{brohan2023rt,kim2024openvla,liu2026plawvlapredictivelatentworld,intelligence2025pi,ma2026internvla}. Track4Action~\cite{wang2026track4actiondistillingworldcentric3d} distills action-aligned 3D transition features into learnable queries that condition action generation. Extending this paradigm, World Action Models (WAMs) ~\cite{yuan2026fast,lou2026dynamicwam,yu2026maskwam,zhang2026imagewamworldactionmodels,wang2026openwamopenmodularexploration} couple future scene prediction with action generation to enhance temporal reasoning and robotic control. Recent approaches further explore efficient predictive modeling: Fast-WAM~\cite{yuan2026fast} leverages video modeling during training without test-time future generation, while ImageWAM~\cite{zhang2026imagewamworldactionmodels} leverages pretrained image-editing models to capture task-relevant
visual transformations without future-video decoding. These advances demonstrate the benefits of predictive world representations for action modeling, while primarily focusing on scene evolution rather than explicit multi-horizon task-progress prediction. 

\begin{figure}[t!]
  \centering
  \includegraphics[width=\textwidth]{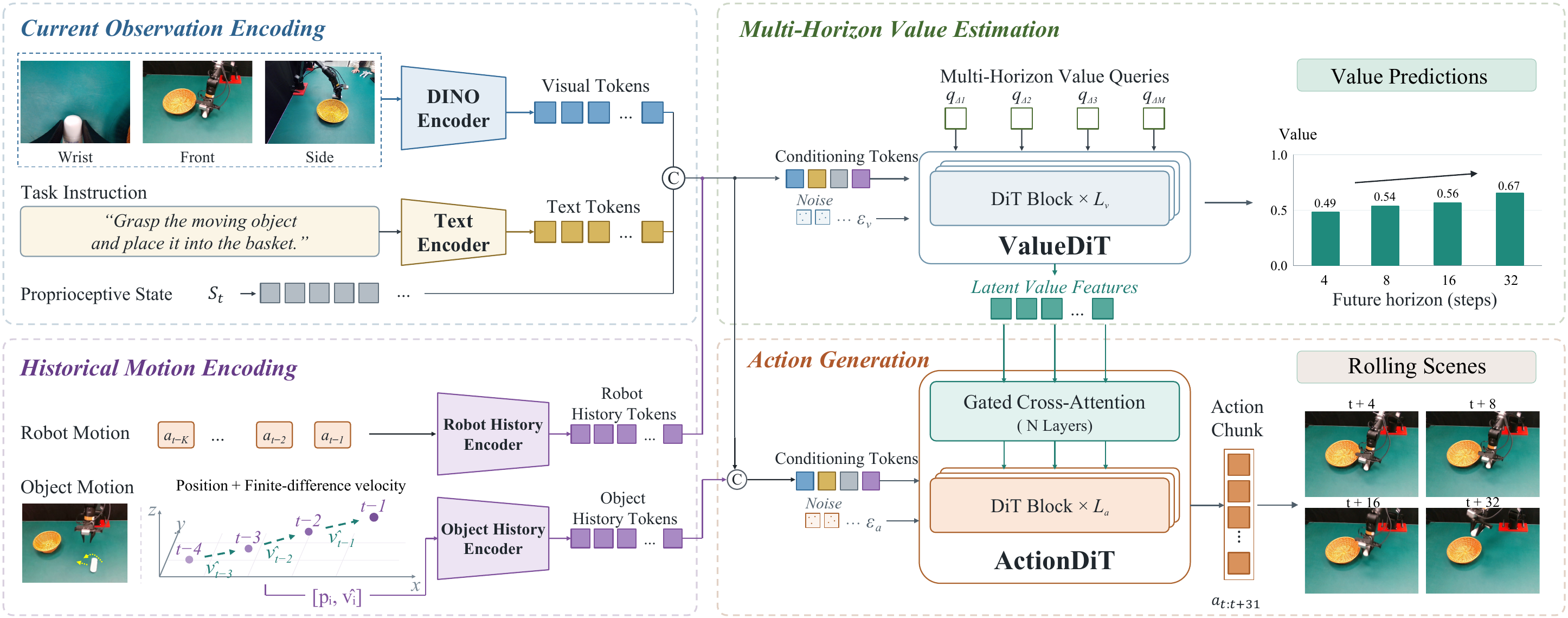}
  \caption{\textbf{Framework of proposed PhysVAM.} Historical object and robot motion provide complementary temporal context, while ValueDiT learns multi-horizon, task-directed future representations. These latent features guide ActionDiT through gated cross-attention to guide action generation toward task success.}
\label{fig:physvam_framework}
  \label{fig:framework}
\end{figure}

\noindent\textbf{Value Models for Robot Learning.}~
Process value learning ~\cite{wang2026world,chen2026robot,fei2026wcm,
intelligence2025pi} enables robots to assess task progress and leverage interaction experience for policy improvement.
Recent studies have advanced value estimation through experience-based learning and predictive world representations.
Specifically, $\pi_{0.6}^{*}$~\cite{intelligence2025pi}
employs value-based advantage learning for policy refinement, while ViVa~\cite{lv2026viva} combines future robot motion prediction with current-state value estimation. WCM~\cite{fei2026wcm} strengthens value estimation through predictive world representations, and
RISE~\cite{yang2026rise} evaluates imagined future states using a current progress value model for policy optimization. WVM~\cite{wang2026world} models value sequences over observed video histories using a coupled video--value architecture. Together, these methods demonstrate the benefits of predictive representations for task-progress assessment. In contrast, PhysVAM focuses on directly predicting task progress at multiple future offsets and incorporating the corresponding query representations into action generation.

\noindent\textbf{Dynamic Object Manipulation.}~ Dynamic Object manipulation requires robots to perceive moving targets, anticipate future interactions, and adapt actions to continuously evolving
physical environments. Recent studies have advanced this direction
through motion-aware perception, predictive planning, and responsive control~\cite{yu2026dynamoma,li2026dsdyn,
park2026dynamic,yun2026physmani}. Among representative methods,
DynamicVLA~\cite{xie2026dynamicvla} improves control responsiveness through continuous inference and latent-aware action streaming.
PUMA~\cite{fang2026towards} exploits historical optical flow and future object-state prediction to enhance motion awareness, while
DynamicWAM~\cite{lou2026dynamicwam} combines historical flow and kinematic motion cues with responsive world-action modeling.
Together, these methods strengthen motion understanding, future anticipation, and timely action execution in dynamic environments. Moreover, PhysVAM utilizes historical physical motion with multi-horizon future value guidance, thus enabling action generation to leverage both target dynamics and task-directed future information.

\section{Method}
\label{sec:method}

\subsection{Overview}
\label{sec:overview}

As illustrated in Fig.~\ref{fig:framework}, PhysVAM addresses dynamic robot manipulation, where objects continue to move during robot
perception and action execution. At time step $t$, the policy takes a language instruction $\ell$,
the current proprioceptive state $s_t$, tri-view visual
observations $O_t$, and histories of object motion $H_t^o$
and robot actions $H_t^a$ to predict an action chunk:
\begin{equation}
    A_t = \pi_{\theta,\phi}
    (O_t,\ell,s_t,H_t^o,H_t^a),
    \quad
    A_t = a_{t:t+S_a-1},
\end{equation}
where $S_a=32$ is the action prediction horizon.
PhysVAM comprises current-observation and historical-motion
encoding modules, together with two coupled branches for
action generation and future value learning.
The action branch employs a 1B-parameter Diffusion
Transformer, termed ActionDiT, whose backbone is initialized using adapted pretrained weights from the Wan2.2-TI2V-5B video DiT~\cite{wan2025wan}, following Fast-WAM~\cite{yuan2026fast}. The value branch employs a smaller Diffusion Transformer, termed ValueDiT, to learn future task-progress representations. Both branches are conditioned on current visual, language,
and proprioceptive features. Historical object-motion features directly condition ActionDiT, while recent robot-action features provide ValueDiT with temporal information about the robot's control behavior. The learned horizon-specific value-query features are integrated into ActionDiT through gated cross-attention to guide action generation toward task success.

\subsection{Historical Motion Encoding}
\label{sec:history}

Unlike static manipulation, where current observations are often sufficient for action prediction, dynamic manipulation requires reasoning about object motion, which may not be reliably inferred from a single current observation~\cite{xie2026dynamicvla, lou2026dynamicwam}. We therefore explicitly encode historical object motion and executed robot actions to provide compact temporal context for action generation and value estimation.

\noindent\textbf{Object Motion.}~
In dynamic environments, task-relevant objects are often in motion, while the surrounding background remains largely static. Historical object motion (e.g., trajectories) can provide compact cues about evolving object dynamics, offering useful temporal context for dynamic manipulation~\cite{lou2026dynamicwam,xie2026dynamicvla}. Motivated by~\cite{zhou2026physicsawareobjectcentricse3forecasting,xie2026dynamicvla, yun2026physmani}, we leverage 3D object positions from four preceding time steps: $P_t^o=[p_{t-4},\ldots,p_{t-1}]$, where $p_i\in\mathbb{R}^3$ to represent historical object trajectories. To further characterize object motion, we estimate velocities using second-order accurate finite differences, with central differences at intermediate time steps and one-sided differences at the temporal boundaries. The resulting position and velocity features are concatenated into 6D object motion states and encoded into object history tokens to condition ActionDiT.

\noindent\textbf{Robot Motion.}~
Robot--object interactions are central to manipulation, as task progress depends on both object dynamics and the robot's control behavior~\cite{lin2026chronoflowpolicyunifyingpastcurrentfutureinteraction, fang2026towards}. In dynamic scenes, historical robot actions reflect the robot's recent control decisions and responses to object motion, providing useful cues for estimating future interactions and task progress~\cite{wang2026remember}. We represent robot motion history using recent robot actions, $H_t^a=\{a_{t-L},\ldots,a_{t-1}\}$, where $L$ denotes the history length. These historical actions are encoded into robot history tokens and provided to ValueDiT, allowing future value estimation to account for the robot's recent control behavior alongside the evolving physical interaction.

\subsection{Multi-Horizon Value-Guided Action Generation}
\label{sec:value_guidance}

Object and robot motion histories provide complementary cues about the evolving robot--object interaction. To anticipate how these interactions will shape future task progress, we introduce a multi-horizon value branch that predicts future task values and learns horizon-specific latent representations to guide action generation toward task success.

\noindent\textbf{Multi-Horizon Value Estimation.}~
To provide dense temporal supervision for future value estimation, we construct temporally structured value targets by combining episode-level outcomes with temporal information. For an episode $e$ of length $T_e$, we assign an outcome label $c(e)=+1$ for success and $c(e)=-1$ for failure. We place $K$ approximately uniformly spaced temporal anchors along each episode:
\begin{equation}
    H_k =
    \max\left(
        1,\left\lceil\frac{kT_e}{K}\right\rceil
    \right),
    \quad k=1,\ldots,K.
\end{equation}
The value target at time step $t\in\{0,\ldots,T_e-1\}$ is defined as:
\begin{equation}
    V_t =
    \frac{c(e)}{K}
    \sum_{k=1}^{K}
    \exp\left(
        -\frac{[H_k-1-t]_+}{\tau}
    \right),
    \label{eq:value_target}
\end{equation}
where $H_k$ denotes the $k$-th temporal anchor, $[x]_+=\max(x,0)$, $\tau>0$ controls the temporal decay, and $K=4$. The episode outcome determines the sign of
$V_t$, while the temporal anchors shape its magnitude over the episode through the exponential terms. The resulting targets lie in $[-1,1]$ and satisfy
$V_{T_e-1}=c(e)$, providing temporally structured supervision that combines task outcomes with temporal progression.

To explicitly supervise future value estimation across the action chunk, we construct targets at multiple future horizons $\mathcal{H}=\{4,8,16,32\}$:
\begin{equation}
    \mathbf{V}_t^{\mathcal{H}} =
    \left[
        V_{\min(t+h,T_e-1)}
    \right]_{h\in\mathcal{H}},
    \label{eq:multi_horizon_value}
\end{equation}
where horizons extending beyond episode termination retain the terminal value. These geometrically spaced horizons provide denser supervision at shorter offsets while covering the full action-chunk length of $S_a=32$ steps. Predicting values at distinct future offsets encourages the model to learn horizon-specific representations of anticipated task progress at multiple temporal scales.

\begin{table}[t!]
    \centering
    \caption{
        \textbf{Main comparison results and transferability of proposed value guidance}. ``+'' denotes augmentation with our value design.
        Best and second-best results are shown in
        \textbf{bold} and \underline{underlined}, respectively.
    }
    \label{tab:main_results}

    \begingroup
    \large
    \setlength{\tabcolsep}{1pt}
    \renewcommand{\arraystretch}{1.25}

    \resizebox{\textwidth}{!}{
        \begin{tabular}{@{}l*{12}{c}@{}}
            \toprule

            \multirow{2}{*}{Methods}
            & \multicolumn{5}{c}{Moderate-speed}
            & \multicolumn{5}{c}{High-speed}
            & \multicolumn{2}{c}{Average} \\

            \cmidrule(lr){2-6}
            \cmidrule(lr){7-11}
            \cmidrule(l){12-13}

            & Avocado
            & Beer
            & Potato
            & Apple
            & Lemon
            & Can
            & D-Bottle
            & W-Bottle
            & Egg
            & Cup
            & SR (\%) $\uparrow$
            & Time (s) $\downarrow$ \\

            \midrule

            FastWAM~\cite{yuan2026fast}
            & \srval{0.00}{0.00}
            & \srval{0.00}{0.00}
            & \srval{0.00}{0.00}
            & \srval{11.67}{8.50}
            & \srval{0.00}{0.00}
            & \srval{0.00}{0.00}
            & \srval{0.00}{0.00}
            & \srval{1.67}{2.36}
            & \srval{3.33}{2.36}
            & \srval{0.00}{0.00}
            & \srval{1.67}{0.47}
            & \srval{11.59}{0.16} \\

            \addlinespace[2pt]

            DynamicWAM~\cite{lou2026dynamicwam}
            & \srval{6.67}{6.24}
            & \srval{5.00}{4.08}
            & \srval{0.00}{0.00}
            & \srval{5.00}{0.00}
            & \srval{1.67}{2.36}
            & \srval{0.00}{0.00}
            & \srval{25.00}{4.08}
            & \srval{8.33}{2.36}
            & \srval{1.67}{2.36}
            & \srval{0.00}{0.00}
            & \srval{5.33}{0.62}
            & \srval{10.96}{0.10} \\

            \addlinespace[2pt]

            InternVLA-A1.5~\cite{ma2026internvla}
            & \srval{3.33}{2.36}
            & \srval{0.00}{0.00}
            & \srval{0.00}{0.00}
            & \srval{3.33}{2.36}
            & \srval{0.00}{0.00}
            & \srval{0.00}{0.00}
            & \srval{18.33}{2.36}
            & \srval{0.00}{0.00}
            & \uval{15.00}{0.00}
            & \srval{0.00}{0.00}
            & \srval{4.00}{0.82}
            & \srval{10.82}{0.03} \\

            \addlinespace[2pt]

            DynamicVLA~\cite{xie2026dynamicvla}
            & \srval{23.33}{2.36}
            & \srval{10.00}{7.07}
            & \srval{5.00}{4.08}
            & \srval{8.33}{2.36}
            & \srval{16.67}{2.36}
            & \uval{1.67}{2.36}
            & \srval{83.33}{10.27}
            & \srval{10.00}{7.07}
            & \srval{6.67}{6.24}
            & \srval{8.33}{8.50}
            & \srval{17.33}{0.24}
            & \srval{8.92}{0.09} \\

            \addlinespace[2pt]

            PUMA~\cite{fang2026towards}
            & \srval{10.00}{0.00}
            & \uval{65.00}{8.16}
            & \srval{41.67}{10.27}
            & \srval{11.67}{8.50}
            & \uval{85.00}{7.07}
            & \uval{1.67}{2.36}
            & \srval{58.33}{2.36}
            & \srval{10.00}{7.07}
            & \srval{0.00}{0.00}
            & \uval{35.00}{4.08}
            & \srval{31.83}{1.70}
            & \srval{9.56}{0.15} \\

            \midrule

            DynamicVLA$^{+}$~\cite{xie2026dynamicvla}
            & \uval{33.33}{6.24}
            & \srval{21.67}{2.36}
            & \srval{8.33}{6.24}
            & \srval{16.67}{2.36}
            & \srval{43.33}{2.36}
            & \srval{0.00}{0.00}
            & \srval{81.67}{6.24}
            & \srval{28.33}{9.43}
            & \srval{13.33}{4.71}
            & \srval{5.00}{0.00}
            & \srval{25.17}{0.85}
            & \srval{9.30}{0.14} \\

            \addlinespace[2pt]

            PUMA$^{+}$~\cite{fang2026towards}
            & \srval{13.33}{4.71}
            & \bval{85.00}{12.25}
            & \uval{51.67}{2.36}
            & \uval{26.67}{6.24}
            & \srval{70.00}{10.80}
            & \srval{0.00}{0.00}
            & \bval{88.33}{2.36}
            & \uval{43.33}{2.36}
            & \srval{5.00}{4.08}
            & \srval{31.67}{31.71}
            & \uval{41.50}{2.86}
            & \uval{8.83}{0.12} \\

            \midrule

            \textbf{PhysVAM (Ours)}
            & \bval{60.00}{10.80}
            & \srval{41.67}{10.27}
            & \bval{65.00}{7.07}
            & \bval{45.00}{4.08}
            & \bval{88.33}{6.24}
            & \bval{36.67}{10.27}
            & \uval{86.67}{8.50}
            & \bval{93.33}{4.71}
            & \bval{28.33}{8.50}
            & \bval{43.33}{4.71}
            & \bval{58.83}{1.65}
            & \bval{8.23}{0.13} \\

            \bottomrule
        \end{tabular}
    }
    \endgroup
\end{table}

\noindent\textbf{Multi-Horizon Value Queries.}~
Conditioned on current observations and robot motion history, ValueDiT employs horizon-specific learnable queries $\{q_h\}_{h\in\mathcal{H}}$ to capture future task progress. The resulting query representations serve both as horizon-specific features for scalar value prediction and as latent conditioning signals for value-guided action generation. The predictions are supervised by:
\begin{equation}
    \mathcal{L}_{\mathrm{reg}} =
    \frac{1}{|\mathcal{H}|}
    \sum_{h\in\mathcal{H}}
    \left(
        \hat{V}_t^{(h)}
        -
        V_{\min(t+h,T_e-1)}
    \right)^2,
    \label{eq:value_regression}
\end{equation}
where $\hat{V}_t^{(h)}$ denotes the prediction made at time $t$ for the future offset $h$. In parallel, ValueDiT learns a flow field that transforms noisy value variables toward multi-horizon value targets. The overall value learning objective combines this loss with horizon-specific query regression:
\begin{equation}
    \mathcal{L}_{\mathrm{value}} =
    \mathcal{L}_{\mathrm{FM}}^{V}
    +
    \mathcal{L}_{\mathrm{reg}}.
    \label{eq:value_loss}
\end{equation}
At inference, future values are predicted from observed interaction context without access to episode-level labels.

\noindent\textbf{Value-Guided Action Generation.}~
To leverage anticipated task progress for action generation,
we use the latent query features as conditioning information
for ActionDiT. Inspired by selective intermediate-layer future conditioning
~\cite{zhao2026faster}, We project horizon-specific features from selected ValueDiT layers into designated ActionDiT layers and integrate them through gated cross-attention:

\begin{equation}
    X_j^A \leftarrow X_j^A +
    G_j \odot \operatorname{CrossAttn}_j
    \left(X_j^A, M_j^V\right),
    \label{eq:value_guided_attention}
\end{equation}
where $X_j^A$ denotes the ActionDiT hidden states at the $j$-th
injection layer, $M_j^V$ contains the projected hidden features of
the horizon-specific value queries, and $G_j$ is a learned,
content-adaptive gate applied element-wise. Each injection layer uses its own gating module. This allows ActionDiT to access representations of anticipated progress at multiple future horizons throughout action generation.

\noindent\textbf{Training Objectives.}~
We jointly train ActionDiT and ValueDiT with:
\begin{equation}
    \mathcal{L} =
    \mathcal{L}_{\mathrm{action}}
    +
    \lambda_V \mathcal{L}_{\mathrm{value}},
    \label{eq:total_loss}
\end{equation}
where $\mathcal{L}_{\mathrm{action}}$ is the action
flow-matching loss and $\lambda_V$ is set to balance action generation
and value learning. Successful trajectories supervise both objectives, whereas
failed trajectories contribute only to the value objective.

\section{Experiments}
\label{sec:experiemnts}

\subsection{Experimental Setup}
\noindent\textbf{Simulation Benchmark.}
Building upon the closed-loop reactivity setting of DOM~\cite{xie2026dynamicvla}, we reconstruct dynamic manipulation scenarios with ten geometrically diverse objects, spanning spherical, ellipsoidal, irregular, cylindrical, bottle-shaped, and cup-shaped forms. To emphasize high-speed interactions, we redesign the object velocity distribution: Can, D-Bottle, W-Bottle, Egg, and Cup scenes are evaluated at $v \geq 0.4\,\mathrm{m/s}$, while the remaining scenes are evaluated at $0.2 \leq v < 0.4\,\mathrm{m/s}$. The resulting dataset comprises 12K successful trajectories and 10K failed trajectories, with the latter collected from rollouts of PUMA~\cite{fang2026towards}, DynamicVLA~\cite{xie2026dynamicvla}, and FastWAM~\cite{yuan2026fast}. Each scene contains at least 1K success and 1K failure trajectories. Following ~\cite{xie2026dynamicvla}, we report task success rate and task completion time. Task completion time is averaged over all trials and include termination by success, timeout, or object drop.

\noindent\textbf{Implementation Details.} 
The ActionDiT comprises 30 transformer layers with a hidden dimension of 1,024, while ValueDiT contains 12 layers with a hidden dimension of 512. Value-query features from ValueDiT layers 4, 8, and 12 are projected and injected into ActionDiT layers 9, 19, and 27, respectively. Both models are jointly trained for 24K steps using AdamW with a learning rate of $10^{-4}$ and a global batch size of 128. The weight $\lambda_V$ is set to 0.1 and history length $L$ is set as four. We set 10 and 4 denoising steps for ActionDiT and ValueDiT, respectively. \textit{More Details are provided in the Supplementary Material.}

\noindent\textbf{Compared Methods.} In simulation, PhysVAM is compared with FastWAM~\cite{yuan2026fast}, InternVLA-A1.5~\cite{ma2026internvla}, DynamicWAM~\cite{lou2026dynamicwam}, DynamicVLA~\cite{xie2026dynamicvla}, and PUMA ~\cite{fang2026towards}. Following~\cite{xie2026dynamicvla}, each method is evaluated over 20 trials for each scene. All methods are fine-tuned using their official codes and released pretrained weights. Results are reported as mean $\pm$ standard deviation over three random seeds.

\subsection{Main Results}
\label{subsec:main_results}

As shown in Table~\ref{tab:main_results}, PhysVAM achieves an average success rate of $58.83\%$, outperforming the strongest original baseline, PUMA, by $27.00$ percentage points. It achieves the highest success rate in eight of ten scenes and the shortest average completion time of $8.23$ seconds under the DOM evaluation protocol. FastWAM and InternVLA-A1.5 achieve average success rates below $6\%$, while DynamicWAM remains limited despite incorporating historical optical flow and kinematic cues, highlighting the difficulty of translating motion awareness into successful interactions. In high-speed scenes, DynamicVLA performs well on D-Bottle but remains inconsistent across other scenes, while DynamicVLA and PUMA achieve only $22.00\%$ and $21.00\%$ success rates, respectively. In contrast, PhysVAM reaches $57.67\%$, demonstrating the benefit of combining historical motion context with future-value guidance for dynamic action generation.

\begin{table}[!htbp]
    \centering
    \caption{
        \textbf{Ablations of value-guided action generation.}
        We examine the value branch, conditioning representations,
        horizon-specific query targets, and future progress
        prediction.
    }
    \label{tab:ablation_value}

    \begingroup
    \fontsize{10}{12}\selectfont
    \renewcommand{\srval}[2]{\mbox{#1{\fontsize{8}{9}\selectfont$\pm$#2}}}
    \setlength{\tabcolsep}{3pt}
    \renewcommand{\arraystretch}{1.15}

    \begin{tabularx}{\columnwidth}{
        @{}>{\raggedright\arraybackslash}Xcc@{}
    }
        \toprule
        \textbf{Model Variant}
        & \textbf{SR (\%) $\uparrow$}
        & \textbf{Time (s) $\downarrow$} \\
        \midrule

        \multicolumn{3}{@{}l@{}}{
            \textit{(a) Value branch}
        } \\

        Without Value Branch
        & \srval{50.50}{1.08}
        & \srval{9.65}{0.02} \\

        \midrule

        \multicolumn{3}{@{}l@{}}{
            \textit{(b) Value representations}
        } \\

        Scalar Conditioning
        & \srval{40.83}{1.43}
        & \srval{9.02}{0.16} \\

        Binary Conditioning
        & \srval{48.50}{2.55}
        & \srval{8.85}{0.08} \\

        \midrule

        \multicolumn{3}{@{}l@{}}{
            \textit{(c) Multi-horizon value prediction}
        } \\

Single-Value (Single Query)
& \srval{37.83}{3.30}
& \srval{9.30}{0.10} \\

Single-Value (Multiple Queries)
& \srval{50.50}{0.71}
& \srval{8.40}{0.13} \\

        \midrule

        \multicolumn{3}{@{}l@{}}{
            \textit{(d) Future progress prediction}
        } \\

        Current-Value Prediction
        & \srval{46.50}{2.27}
        & \srval{9.02}{0.16} \\

        \midrule

        \textbf{PhysVAM (Full)}
        & \textbf{\srval{58.83}{1.65}}
        & \textbf{\srval{8.23}{0.13}} \\

        \bottomrule
    \end{tabularx}
    \endgroup
\end{table}

\begin{table}[!htbp]
    \centering
    \caption{
        \textbf{Ablations of object and robot motion histories}. All historical inputs span four frames.
    }
    \label{tab:ablation_motion}

    \begingroup
    \fontsize{10}{12}\selectfont
    \renewcommand{\srval}[2]{\mbox{#1{\fontsize{8}{9}\selectfont$\pm$#2}}}
    \setlength{\tabcolsep}{3pt}
    \renewcommand{\arraystretch}{1.15}

    \begin{tabularx}{\columnwidth}{
        @{}>{\raggedright\arraybackslash}Xcc@{}
    }
        \toprule
        \textbf{Model Variant}
        & \textbf{SR (\%) $\uparrow$}
        & \textbf{Time (s) $\downarrow$} \\
        \midrule

        \multicolumn{3}{@{}l@{}}{
            \textit{(a) Object motion representation}
        } \\

        Without Object Motion History
        & \srval{17.17}{2.39}
        & \srval{9.65}{0.02} \\

        RGB History
        & \srval{22.33}{1.55}
        & \srval{9.27}{0.26} \\

        Object Position History
        & \srval{30.67}{2.01}
        & \srval{9.53}{0.03} \\

        Object Velocity History (Oracle)
        & \srval{34.00}{3.27}
        & \srval{9.00}{0.10} \\

        Object Velocity History (Estimated)
        & \srval{33.67}{0.85}
        & \srval{8.93}{0.04} \\

        \midrule

        \multicolumn{3}{@{}l@{}}{
            \textit{(b) Robot motion history}
        } \\

        Without Robot Motion History
        & \srval{45.17}{2.05}
        & \srval{8.87}{0.04} \\

        \midrule

        \textbf{PhysVAM (Full)}
        & \textbf{\srval{58.83}{1.65}}
        & \textbf{\srval{8.23}{0.13}} \\

        \bottomrule
    \end{tabularx}
    \endgroup
\end{table}

The benefits of future-value guidance also extend to other action models. Integrating our value design improves the average success rates of DynamicVLA and PUMA from $17.33\%$ to $25.17\%$ and from $31.83\%$ to $41.50\%$, respectively, demonstrating its effectiveness across different action policies.

\subsection{Ablation Studies}
\label{subsec:ablation}

\noindent\textbf{Effect of the Value Branch.}
We evaluate the contribution of future-value guidance to action generation. As shown in Table~\ref{tab:ablation_value}, removing the value branch reduces the success rate from $58.83\%$ to $50.50\%$, demonstrating the effectiveness of future-value guidance for dynamic manipulation.

\noindent\textbf{Effect of Value Representations.}
We replace the latent value representations injected into the action policy with embeddings of either four predicted scalar values or binary good/bad indicators obtained by mapping these predictions to $\{0,1\}$, following~\cite{intelligence2025pi}. As shown in Table~\ref{tab:ablation_value}, both scalar and binary conditioning underperform latent conditioning, suggesting that latent value representations capture richer task-relevant information than explicit value scores and provide more effective action guidance.

\noindent\textbf{Effect of Multi-Horizon Value Prediction.}
We vary the number of value queries and their supervision targets while keeping the future-value flow-matching objective unchanged. Under a shared overall future-value target (averaged by four targets), increasing the number of queries from one to four improves the success rate from $37.83\%$ to $50.50\%$. Assigning distinct horizon-specific targets to the four queries further improves performance to $58.83\%$. These results indicate that multiple queries benefit value learning even under shared supervision, while horizon-specific targets provide additional gains by capturing task progress across different future horizons.

\begin{figure}[!t]
  \centering
  \includegraphics[width=0.82\columnwidth]{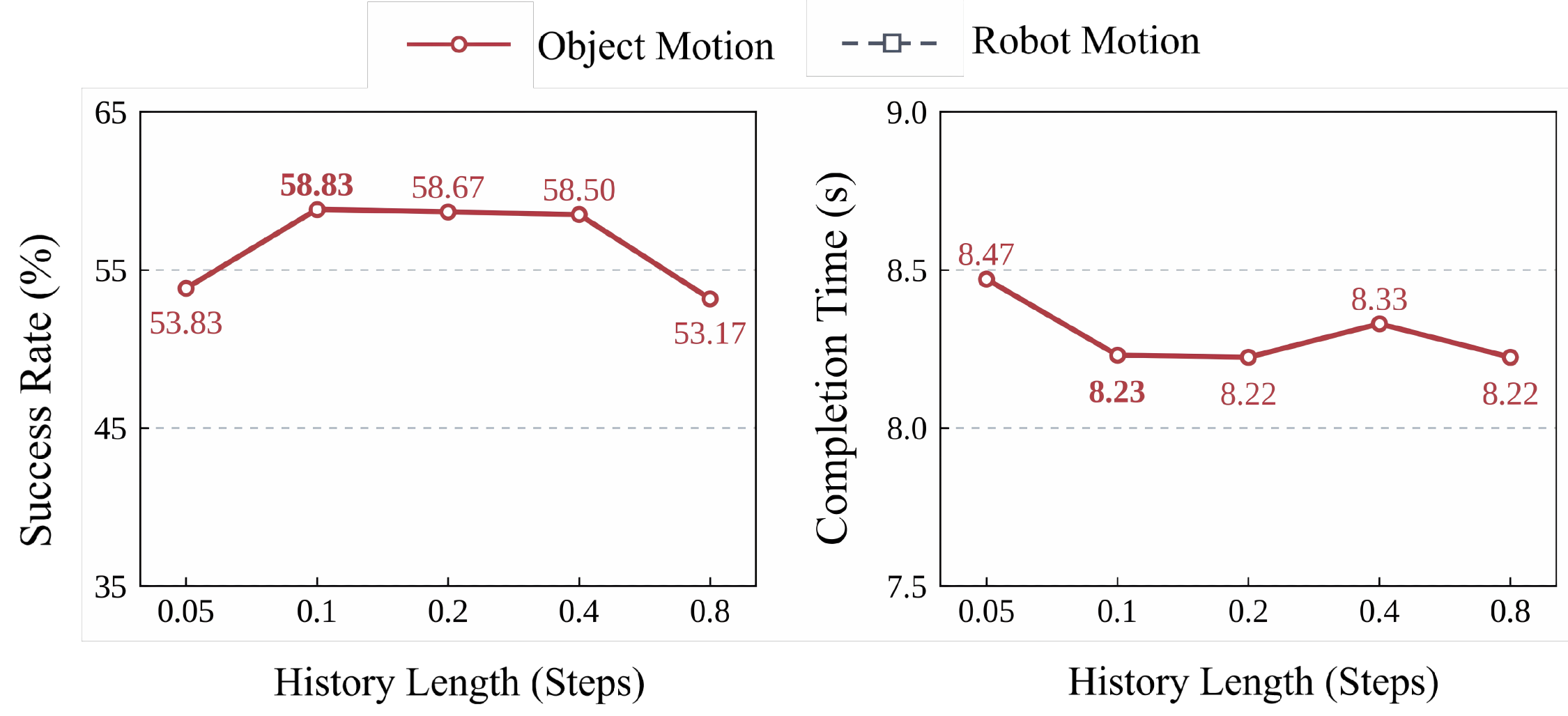}
  \caption{\textbf{Effect of Object and Robot Motion History Length.}}
  \label{fig:history_length}
\end{figure}

\noindent\textbf{Effect of Future Progress Prediction.}
We replace the future-offset query targets with current-value targets ($h=0$), keeping the progress definition, query count, and injection scheme unchanged. This reduces the success rate from $58.83\%$ to $46.50\%$, a $12.33$-percentage-point drop. Current-value supervision focuses on present progress, whereas future-offset targets encourage horizon-specific representations of how upcoming interactions may advance or hinder the task. These results support our value-based future modeling formulation, highlighting the benefit of anticipating task progression for dynamic action generation.

\begin{figure}[h]
  \centering
  \includegraphics[width=\textwidth]{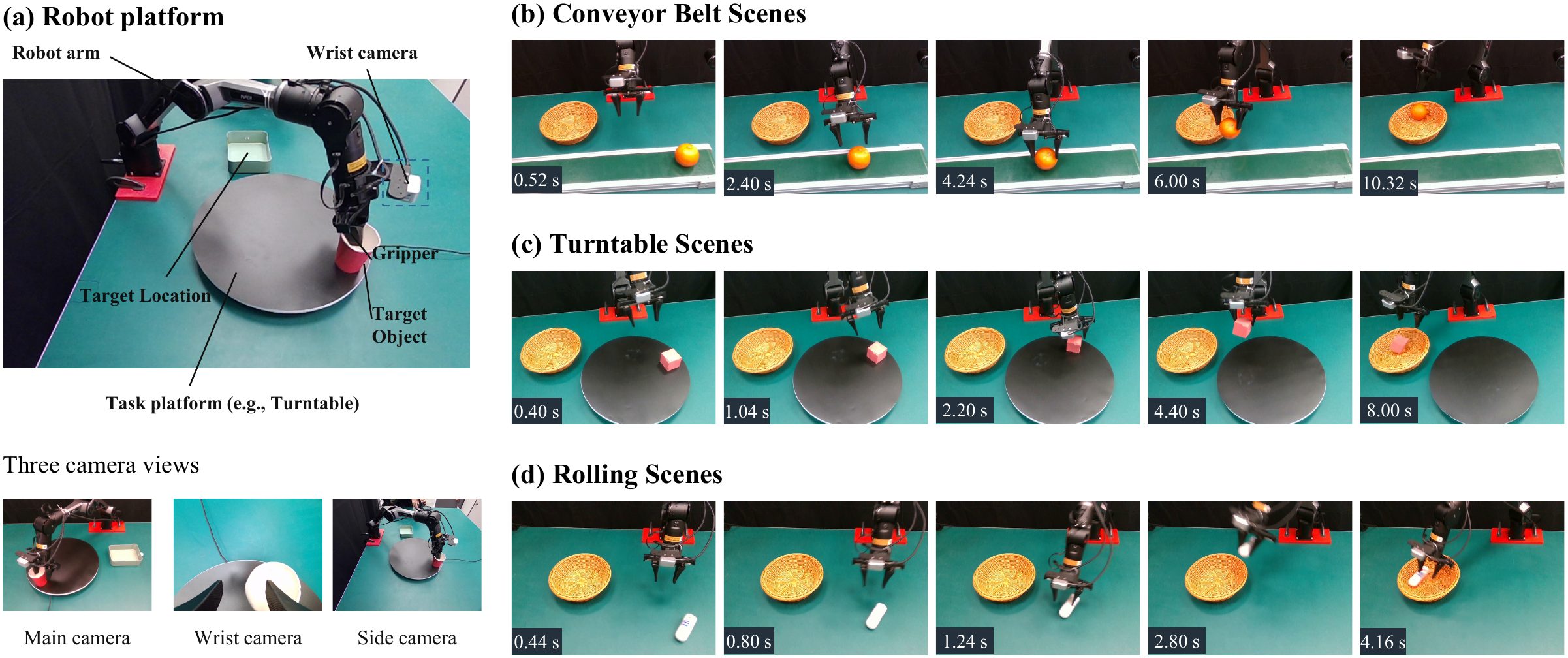}
  \caption{\textbf{Real-world experimental setup}. Our platform uses a single AgileX PiPER arm and includes multiple targets that undergo linear, circular, and compound rolling dynamic manipulation tasks.}
  \label{fig:real_world}
\end{figure}

\noindent\textbf{Interaction-Sensitive Future-Value Predictions.}~
Beyond task performance, we examine whether future-value predictions reflect interaction-dependent changes in task progress. We focus on the shortest prediction horizon ($h=4$) and evaluate value changes over a $\pm16$-step window centered on gripper closure. Across 220 grasp attempts from 200 simulation episodes, predicted values increase by $0.254$ for successful lifts and decrease by $0.185$ for missed grasps on average, yielding a Lift--Miss difference of $0.439$. Predicted values increase in 96/119 lift attempts and decrease in 77/101 missed attempts, consistent with the progress gains and setbacks associated with these interactions. Notably, across 33 episodes containing both interaction outcomes, the mean within-episode Lift--Miss difference remains positive ($+0.319$), further supporting sensitivity to local interactions beyond episode-level outcome labels. These results provide quantitative evidence that future-value predictions respond to evolving robot--object interactions. \textit{Further analyses and more visualizations are provided in the Supplementary Material.}

\noindent\textbf{Effect of Object Motion History.}
We compare different object-motion representations in Table~\ref{tab:ablation_motion}(a), while keeping the robot-action history fixed. Without object-motion history, the success rate drops to $17.17\%$, highlighting the importance of object dynamics for manipulation. Incorporating historical RGB observations or object positions improves performance to $22.33\%$ and $30.67\%$, respectively, indicating the advantage of explicit spatial trajectories over implicit visual histories.
Moreover, position-derived velocity achieves a success rate of $33.67\%$, comparable to $34.00\%$ with oracle velocity, suggesting that effective motion cues can be obtained without privileged velocity information. Combining object positions and estimated velocities further increases performance to $58.83\%$, substantially outperforming either representation alone. This improvement highlights the complementary roles of spatial position and motion velocity in anticipating target displacement and coordinating dynamic grasps.

\noindent\textbf{Effect of Robot Motion History.}
As shown in Table~\ref{tab:ablation_motion}(b), removing only historical robot actions reduces the success rate from $58.83\%$ to $45.17\%$, indicating that object-motion information alone is insufficient for effective dynamic manipulation. Historical robot actions provide complementary temporal context for coordinating robot movements with evolving object dynamics.

\noindent\textbf{Effect of History Length.}
As shown in Fig.~\ref{fig:history_length}, we vary object and robot motion history lengths independently, fixing the other at four frames. Increasing either history from two to four frames improves success from $39.83\%$ to $58.83\%$ and from $43.67\%$ to $58.83\%$, respectively, highlighting the importance of temporal context for object dynamics and robot coordination. Extending either to six frames reduces success to $49.67\%$ and $46.00\%$, respectively, suggesting redundant information. Finlay, we set the length of history frames as four. \textit{Further hyperparameter analyses, temporal action reweighting, and per-object results are provided in the Supplementary Material.}

\subsection{Real-World Experiments}
\label{subsec:real_world}

\paragraph{Experimental Setup.}
As illustrated in Fig.~\ref{fig:real_world}, we evaluate PhysVAM on a single AgileX PiPER robotic arm across three dynamic manipulation scenarios involving distinct motion patterns: conveyor-belt translation
(0.07--0.17 m/s), turntable rotation
(25.33$^\circ$/s), and object rolling
(0.20--0.30 m/s). Our real-world dataset comprises 60 conveyor-belt, 60 turntable, and 30 rolling trajectories involving objects with diverse geometries, including oranges, cuboid blocks, bread, cups, and bottles. We compare PhysVAM with DynamicVLA~\cite{xie2026dynamicvla} and PUMA~\cite{fang2026towards}. All three methods are fine-tuned on the real-world dataset, with the baselines initialized from simulation-fine-tuned checkpoints.

\begin{figure}[h]
  \centering
  \includegraphics[width=0.82\columnwidth]{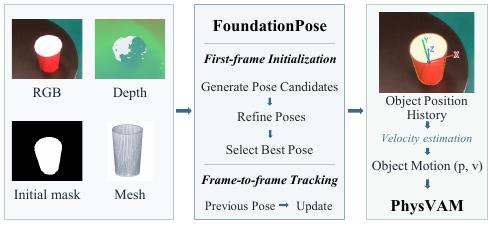}
  \caption{
    \textbf{Object motion estimation for real-world tasks.}
    We employ FoundationPose to estimate and track object poses from RGB-D observations using an initial mask and an object mesh. Four-frame position histories and their derived velocities jointly form the object motion inputs to PhysVAM.
  }
  \label{fig:pose}
\end{figure}

\noindent\textbf{Object Motion Estimation.}
As illustrated in Fig.~\ref{fig:pose}, we estimate object motion from the main-view RGB-D stream using FoundationPose~\cite{wen2024foundationpose}, with color-based foreground segmentation and an object mesh generated by OnePoseviaGen~\cite{geng2025one}. In the first frame, FoundationPose generates and refines pose candidates to select an initial 3D object pose. It then tracks the object by refining the previous pose against subsequent RGB-D observations. We extract object positions from the estimated poses and compute historical velocities using the same formulation as in simulation.

\begin{figure}[t]
  \centering
  \includegraphics[width=\linewidth]{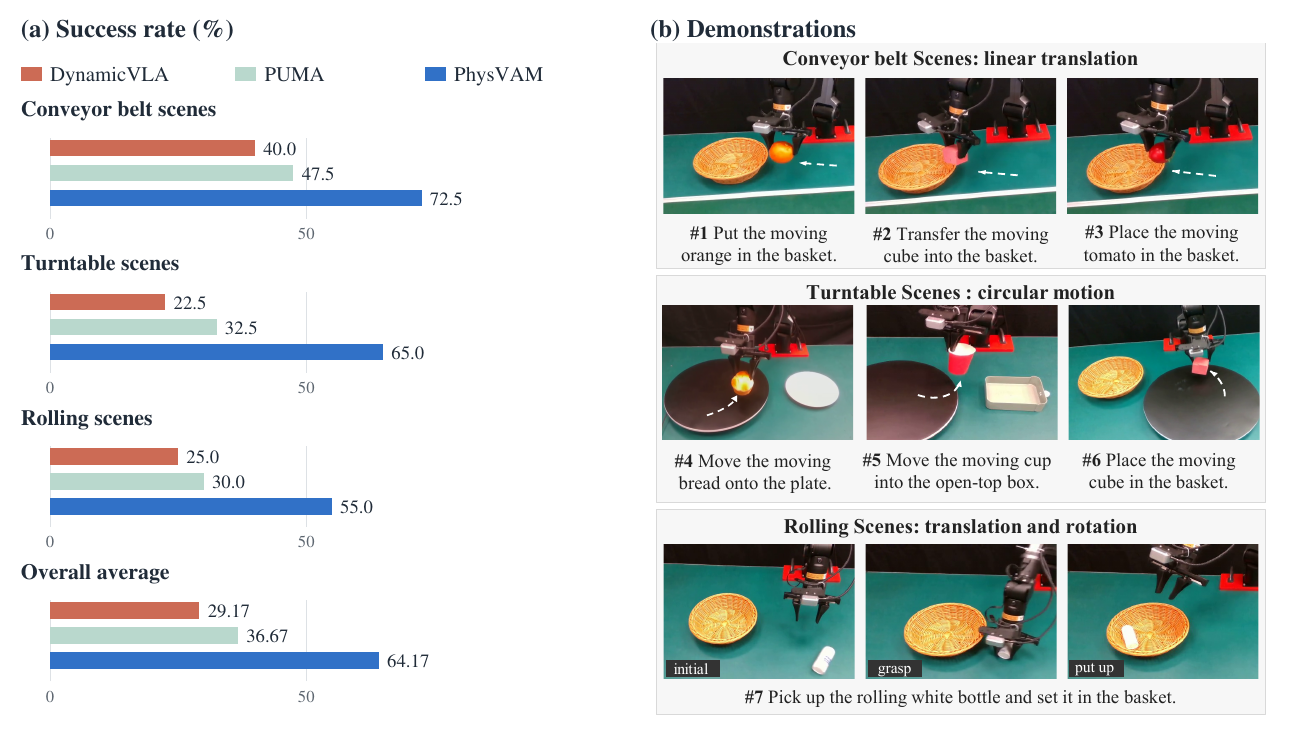}
  \caption{
    \textbf{Real-world experimental results for DynamicVLA, PUMA, and PhysVAM under multiple scenes.}
  }
  \label{fig:dynamic_tasks}
\end{figure}

\noindent\textbf{Results.} Fig.~\ref{fig:dynamic_tasks} presents real-world success rates and representative executions across three dynamic manipulation scenarios. Macro-averaged across three scenarios, PhysVAM achieves a success rate of 64.17\%, outperforming PUMA by 27.50 percentage points. Specifically, PhysVAM achieves $72.5\%$, $65.0\%$, and $55.0\%$ on conveyor belt, turntable, and rolling tasks, exceeding the best compared method by $25.0$, $32.5$, and $25.0$ percentage points, respectively. As illustrated in Fig.~\ref{fig:dynamic_tasks} (b), these scenarios present distinct motion challenges: linear translation requires anticipating target displacement and grasp timing, circular motion involves continuously changing motion directions, and rolling couples translation with object rotation, further challenging grasp alignment. The consistent improvements across diverse objects and motion patterns demonstrate the effectiveness of combining historical motion context with future-value guidance for dynamic action generation. Test Trials are 20 for each object in each scene. \textit{Detailed results are provided in the Supplementary Material.}

\section{Conclusion and Future Work}

In this paper, we propose \textbf{PhysVAM}, a Physics-oriented Future Value Action Model that instantiates value-based future modeling for dynamic manipulation. By integrating historical physical motion modeling with multi-horizon future-value prediction, PhysVAM learns task-directed latent representations to guide action generation toward task success. This formulation characterizes anticipated robot--object interactions through their effects on task progress, providing task-directed foresight without reconstructing future visual states. Extensive simulation and real-world experiments demonstrate the effectiveness of PhysVAM in improving task success and execution efficiency, while gains across other action models further support the transferability of future-value guidance.

Looking ahead, there are three promising directions for future work: 1) Toward Value-Guided Self-Improvement. Our framework relies on offline trajectory supervision.
Future work could leverage future value predictions to identify interaction setbacks, guide recovery and exploration, and provide progress-aware feedback for online reinforcement learning. This could enable iterative policy improvement through continuous interaction. 2) More Complex Dynamic Manipulation. Future work could explore more object motions, multi-object interactions, and longer-horizon tasks requiring sustained manipulation. 3) Value-Guided World-Action Models. PhysVAM predicts future task values without explicitly modeling future scene dynamics. Integrating multi-horizon value guidance with predictive WAMs could combine scene-dynamics prediction with task-directed action generation, while maintaining efficient inference for real-time manipulation.

\clearpage

\clearpage
\beginappendix
\section{Method}

\subsection{Training Algorithm}

Algorithm~\ref{alg:physvam} summarizes the joint training of ValueDiT and ActionDiT, where value representations guide action generation through gated cross-attention.
\begin{algorithm}[H]
\small
\caption{Training Pipeline of the Proposed PhysVAM}
\label{alg:physvam}

\renewcommand{\algorithmicrequire}{\textbf{Input:}}
\renewcommand{\algorithmicensure}{\textbf{Output:}}

\setlength{\algorithmicindent}{1.15em}

\begin{algorithmic}[1]
\setlength{\itemsep}{1pt}

\REQUIRE \parbox[t]{0.88\linewidth}{
Training episodes $\mathcal D$ with outcome labels $c(e)$
and precomputed targets $\mathbf V_t^{\mathcal H}$;
future horizons $\mathcal H=\{4,8,16,32\}$;
value loss weight $\lambda_V$.
}

\ENSURE Trained policy $\pi_\theta$ and value model $\phi$

\FOR{each sampled episode--timestep pair $(e,t)$}

    \STATE \textbf{1. Multi-Horizon Future Value Learning}

    \STATE $(\hat{\mathbf V}_t^{\mathcal H},M_t^V)
    \gets \text{ValueDiT}(O_t,H_t^a,\{q_h\}_{h\in\mathcal H})$

    \STATE $\mathcal L_{\text{value}}
    \gets \mathcal L_{\text{FM}}^V+\mathcal L_{\text{reg}}$

    \STATE \textbf{2. Value-Guided Action Generation}

    \STATE $\mathcal L_{\text{action}}\gets 0$

    \IF{$c(e)=+1$}

        \STATE $C_t=(\ell,s_t,O_t,H_t^o,H_t^a)$

        \STATE $\hat A_t
        \gets \text{ActionDiT}(C_t,M_t^V)$

        \STATE $\mathcal L_{\text{action}}
        \gets \text{ActionLoss}(\hat A_t,A_t)$

    \ELSE

        \STATE \textit{Skip action loss for failed trajectories.}

    \ENDIF

    \STATE $\mathcal L
    \gets \mathcal L_{\text{action}}
    +\lambda_V\mathcal L_{\text{value}}$

    \STATE Update $(\theta,\phi)$ using
    $\nabla_{\theta,\phi}\mathcal L$

\ENDFOR

\RETURN $\pi_\theta,\phi$

\end{algorithmic}
\end{algorithm}

\subsection{Implementation Details}
The value loss weight $\lambda_V$ is set to 0.1 and linearly warmed up over the first 2K steps. Action DiT and ValueDiT are jointly trained for 24K steps using AdamW with a learning rate of $10^{-4}$ and a global batch size of 128 on four NVIDIA RTX 5090 GPUs. Training requires approximately 48 GPU-hours. During inference, PhysVAM employs asynchronous action-chunk generation, achieving a latency of 0.322 seconds per chunk on a single NVIDIA RTX 5090 GPU. For $\tau$ in Eq. (3), we set $\tau=16$, corresponding to half the action-chunk length, to provide gradual temporal variation in the value targets while retaining sensitivity to the distance from each temporal anchor.

Inspired by prior work~\cite{lin2026roboflow4d,zhao2026faster},
we condition ActionDiT on horizon-specific value representations.
We extract the hidden states of the value queries from ValueDiT blocks 4, 8, and 12, project them, and inject them into ActionDiT layers 9, 19, and 27, respectively. At each injection layer, the action hidden states attend to the corresponding value features through a gated residual update (Eq.~\eqref{eq:value_guided_attention}). A layer-specific gating network takes the action hidden states, pooled value features, and denoising timestep features as inputs and produces token- and channel-wise gates. The same network predicts the scale and shift applied to the normalized attention queries.

\clearpage
\section{Additional Experiments Results}
\label{sec:Additional Experiments Results}

\begin{table}[t]
    \centering
    \caption{
        \textbf{Closure-aligned future-value changes in simulation.}
        Mean $\pm$ standard deviation across 220 valid grasp attempts from 200 evaluation trajectories, including retries.
    }
    \label{tab:closure_value_analysis}

    \begingroup
    \small
    \setlength{\tabcolsep}{4pt}
    \renewcommand{\arraystretch}{1.2}

    \begin{tabular}{@{}lccc@{}}
        \toprule
        Window
        & \shortstack{Lift\\($n=119$)}
        & \shortstack{Miss\\($n=101$)}
        & Gap \\
        \midrule
        $\pm8$
        & $+0.103\pm0.279$
        & $-0.150\pm0.240$
        & $+0.253$ \\

        $\pm16$
        & $+0.254\pm0.370$
        & $-0.185\pm0.237$
        & $\mathbf{+0.439}$ \\
        \bottomrule
    \end{tabular}
    \endgroup
\end{table}

\subsection{Interaction-Sensitive Future Value Predictions}
\label{subsec:value_interaction_sensitivity}

\noindent\textbf{Quantitative Analysis.}~
We examine whether future-value predictions respond to local interaction outcomes by aligning grasp attempts to gripper closure. We adopt the shortest prediction horizon ($h=4$) to capture near-term interaction responses and evaluate value changes over $\pm8$ ($\pm0.32$ s) and $\pm16$ ($\pm0.64$ s) control steps across 220 valid grasp attempts from 200 simulation trajectories. For a closure event at $t_c$, we calculate:
\begin{equation}
    \Delta\hat V_{\delta}^{(4)} =
    \hat V_{t_{\delta}^{+}}^{(4)}
    - \hat V_{t_{\delta}^{-}}^{(4)},
    \label{eq:closure_value_change}
\end{equation}
where $t_{\delta}^{-}$ and $t_{\delta}^{+}$
denote the nearest valid prediction times
to $t_c-\delta$ and $t_c+\delta$, respectively.
These temporal windows are distinct from
the prediction horizon $h$.

\begin{figure}[t]
    \centering
    \includegraphics[width=0.97\textwidth]{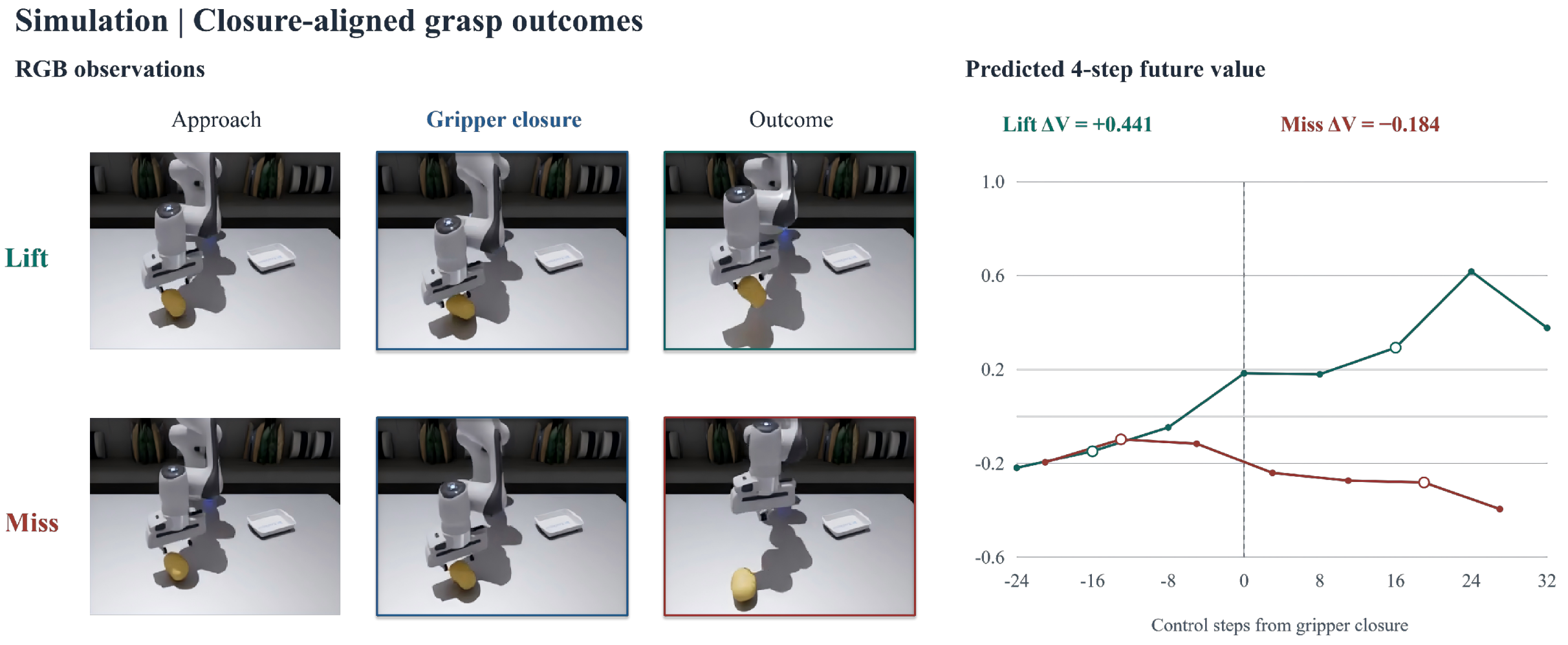}
    \caption{
        \textbf{Closure-aligned future-value predictions
        for Lift and Miss attempts in simulation.}
        Representative attempts are collected in the same
        environment using the same policy, with similar
        object--gripper positions at closure.
        RGB observations and 4-step-ahead task-value predictions
        illustrate the progress gains and setbacks associated
        with the two interaction outcomes.
        The vertical reference line marks gripper closure.
    }
    \label{fig:value_sim_closure}
\end{figure}

As shown in Table~\ref{tab:closure_value_analysis}, future-value predictions reflect local gains and setbacks in task progress. For the $h=4$ predictions within the $\pm16$-step window, successful lifts yield a mean value increase of $0.254$, whereas missed grasps exhibit a decrease of $0.185$, giving a Lift--Miss gap of $0.439$. Predicted values increase in $80.7\%$ of Lift attempts and decrease in $76.2\%$ of Miss attempts. Across 33 episodes containing both interaction types, the mean within-episode Lift--Miss difference remains positive at $+0.319$.

\noindent\textbf{Closure-Aligned Simulation Visualization.}~
Fig.~\ref{fig:value_sim_closure} compares representative Lift and Miss attempts with visually similar object--gripper configurations at closure. As the interactions unfold, the predicted value increases by $0.441$ for the successful lift and decreases by $0.184$ for the missed grasp. These examples illustrate the distinction between an interaction's instantaneous configuration and its subsequent task consequences, which are reflected in the evolving value predictions.

\begin{figure}[t]
    \centering
    \includegraphics[width=0.97\textwidth]{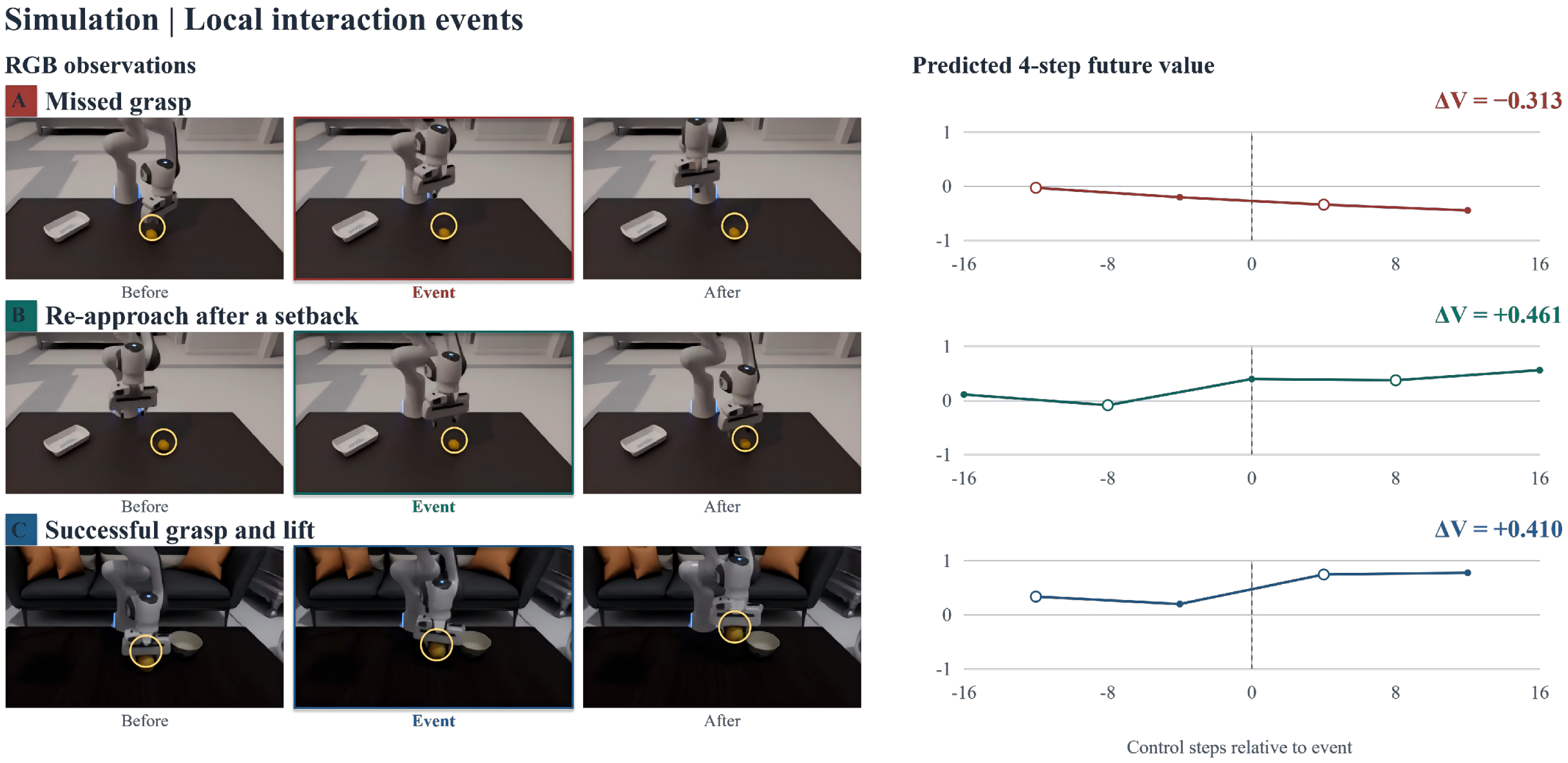}
    \caption{
        \textbf{Interaction-sensitive future-value
        predictions in simulation.}
        RGB observations and 4-step-ahead task-value predictions
        illustrate (a) a missed grasp,
        (b) re-approach after a setback, and
        (c) successful grasping and lifting.
        Each row shows a different trajectory.
        Dashed lines mark the corresponding interaction events,
        and yellow circles highlight the target objects.
    }
    \label{fig:value_sim_events}
\end{figure}

\noindent\textbf{Additional Simulation Interactions.}~
Fig.~\ref{fig:value_sim_events} presents representative interaction events from different trajectories, including missed grasps, re-approaching after setbacks, and successful grasping and lifting. Predicted values decrease around missed grasps and increase during re-approaching and successful lifting. These examples illustrate how future-value predictions respond to distinct gains and setbacks in robot--object interactions.

\begin{figure}[t]
    \centering
    \includegraphics[width=0.97\textwidth]{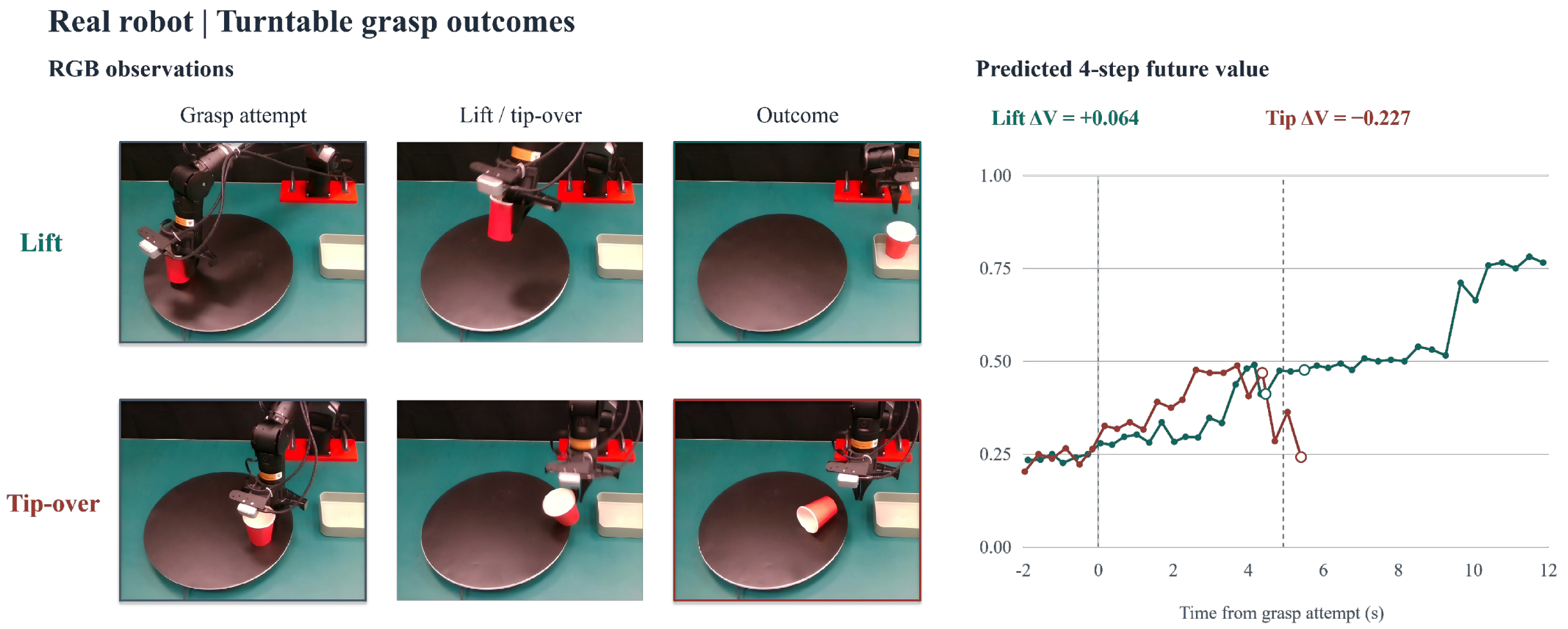}
    \caption{
        \textbf{Real-world future-value predictions
        during turntable manipulation.}
        RGB observations and 4-step-ahead task-value predictions
        illustrate progress gains during successful lifting
        and a setback when the cup tips over.
        Reported value changes are measured around the
        corresponding lift or tip-over event.
    }
    \label{fig:value_real_turntable}
\end{figure}

\noindent\textbf{Real-World Interaction Analysis.}~
We further examine future-value predictions in real-world turntable manipulation, where object motion continues during execution. As illustrated in Fig.~\ref{fig:value_real_turntable}, the predicted value increases by $0.064$ during successful lifting and decreases by $0.227$ when the cup tips over. These examples provide qualitative evidence of interaction-sensitive value changes during real physical execution. The response to cup tipping further illustrates how the same progress-based interpretation can describe a different failure mode.

\noindent\textbf{Beyond Temporally Shaped Value Targets.}~
To examine whether local value responses merely follow the prescribed temporal supervision, we compare predicted and target value changes around grasp attempts. Among 28 missed grasps from 25 ultimately successful simulation episodes, predicted values decrease in 19 attempts (67.9\%) over a $\pm16$-step window, whereas the corresponding targets increase in all 28 attempts. For 33 episodes containing both successful lifts and missed grasps, the within-episode Lift--Miss difference is $+0.319$ for predictions but only $+0.043$ for targets, yielding a paired difference of $+0.277$ (95\% bootstrap CI: $[+0.125,+0.431]$). These results indicate that learned future-value predictions exhibit local variations not explicitly encoded by the temporal supervision targets.

Overall, these analyses support our value-based future modeling formulation: predicted future task values describe the gains, setbacks, and recovery associated with evolving robot--object interactions. The value targets combine temporal progression with final episode outcomes, without explicit supervision for individual interaction events. Together with the main-paper ablation comparing current-value and multi-horizon future-value prediction, these findings further support the benefits of future-value learning for dynamic action generation.

\subsection{Value-Guided Action Reweighting}

To investigate whether ValueDiT can provide reward-like signals for action learning, we use the best-performing checkpoint as a frozen teacher to evaluate task progress in offline datasets. Specifically, we predict multi-horizon values at 8-frame intervals and derive segment-level progress scores from consecutive $V_4$ predictions. We use an 8-frame interval for offline progress scoring to capture changes over longer action segments. Since higher values indicate greater task progress, we define:
\begin{equation}
    \Delta V_i = V_4(s_{t_i+8})-V_4(s_{t_i}).
\end{equation}
We convert positive progress scores into normalized action weights:
\begin{equation}
    p_i=\max(\Delta V_i,0),
    w_i=\operatorname{clip}\left(
    1+\frac{p_i}{\operatorname{mean}_{j:p_j>0}(p_j)},
    1,4\right).
\end{equation}
Each weight is applied to its corresponding 32-step action chunk, yielding the weighted action objective:
\begin{equation}
    \mathcal L_{\mathrm{action}}
    = \frac{\sum_i w_i c_i \ell_i}{\sum_i w_i}.
\end{equation}
Finally, ActionDiT and ValueDiT are jointly retrained from the same base initialization. As shown in Table~\ref{tab:additional_experiments}, value-guided reweighting improves the average success rate from $58.83\%$ to $61.50\%$, yielding a $2.67$ percentage-point gain. Improvements are observed in six of the ten object categories, with particularly notable gains on the cup, avocado, and beer. These results suggest that ValueDiT can provide informative progress signals that can be transferred to action learning through reward-like reweighting.

\begin{table}[t]
    \centering
    \caption{
        \textbf{Effects of value-guided action reweighting
        and multi-view observations.}
        Main (3-CAM) denotes the default setting,
        Reweighted applies value-guided action weights,
        and Main (2-CAM) uses two camera views.
        Best results are shown in bold.
    }
    \label{tab:additional_experiments}

    \begingroup
    \large
    \setlength{\tabcolsep}{1pt}
    \renewcommand{\arraystretch}{1.25}

    \resizebox{\textwidth}{!}{
        \begin{tabular}{@{}l*{12}{c}@{}}
            \toprule

            \multirow{2}{*}{Methods}
            & \multicolumn{5}{c}{Moderate-speed}
            & \multicolumn{5}{c}{High-speed}
            & \multicolumn{2}{c}{Average} \\

            \cmidrule(lr){2-6}
            \cmidrule(lr){7-11}
            \cmidrule(l){12-13}

            & Avocado
            & Beer
            & Potato
            & Apple
            & Lemon
            & Can
            & D-Bottle
            & W-Bottle
            & Egg
            & Cup
            & SR (\%) $\uparrow$
            & Time (s) $\downarrow$ \\

            \midrule

            Main (3-CAM)
            & \srval{60.00}{10.80}
            & \srval{41.67}{10.27}
            & \textbf{\srval{65.00}{7.07}}
            & \textbf{\srval{45.00}{4.08}}
            & \textbf{\srval{88.33}{6.24}}
            & \srval{36.67}{10.27}
            & \srval{86.67}{8.50}
            & \textbf{\srval{93.33}{4.71}}
            & \srval{28.33}{8.50}
            & \srval{43.33}{4.71}
            & \srval{58.83}{1.65}
            & \textbf{\srval{8.23}{0.13}} \\

            Reweighted (3-CAM)
            & \textbf{\srval{66.00}{4.08}}
            & \textbf{\srval{49.33}{8.50}}
            & \srval{61.00}{4.08}
            & \srval{42.67}{2.36}
            & \srval{84.33}{6.24}
            & \textbf{\srval{37.67}{4.71}}
            & \srval{89.33}{4.71}
            & \srval{91.00}{10.80}
            & \textbf{\srval{29.33}{2.36}}
            & \srval{64.33}{6.24}
            & \textbf{\srval{61.50}{1.08}}
            & \srval{8.88}{0.57} \\

            Main (2-CAM)
            & \srval{55.00}{4.08}
            & \srval{46.67}{10.27}
            & \srval{46.67}{4.71}
            & \srval{25.00}{10.80}
            & \srval{45.00}{4.08}
            & \srval{8.33}{4.71}
            & \textbf{\srval{93.33}{6.24}}
            & \srval{86.67}{6.24}
            & \srval{28.33}{6.24}
            & \textbf{\srval{65.00}{4.08}}
            & \srval{50.00}{2.55}
            & \srval{9.05}{0.10} \\

            \bottomrule
        \end{tabular}
    }
    \endgroup
\end{table}

\subsection{Effect of Multi-View Information}
We further explore the effectiveness of multi-view information, as shown in Table~\ref{tab:additional_experiments}, comparing three-view and two-view configurations (without side view), three-view observations achieves the average success rate $58.83\%$ compared with $50.00\%$, while reducing the average execution time from $9.05$~s to $8.23$~s. The improvements are observed in six
of the ten object categories, with particularly substantial gains on the can, apple, and lemon. This phenomenon indicates that complementary viewpoints can provide additional visual cues for tracking evolving object states and motion, helping the policy better adapt to dynamic targets during manipulation.

\begin{figure}[t!]
  \centering
  \includegraphics[width=0.82\columnwidth]{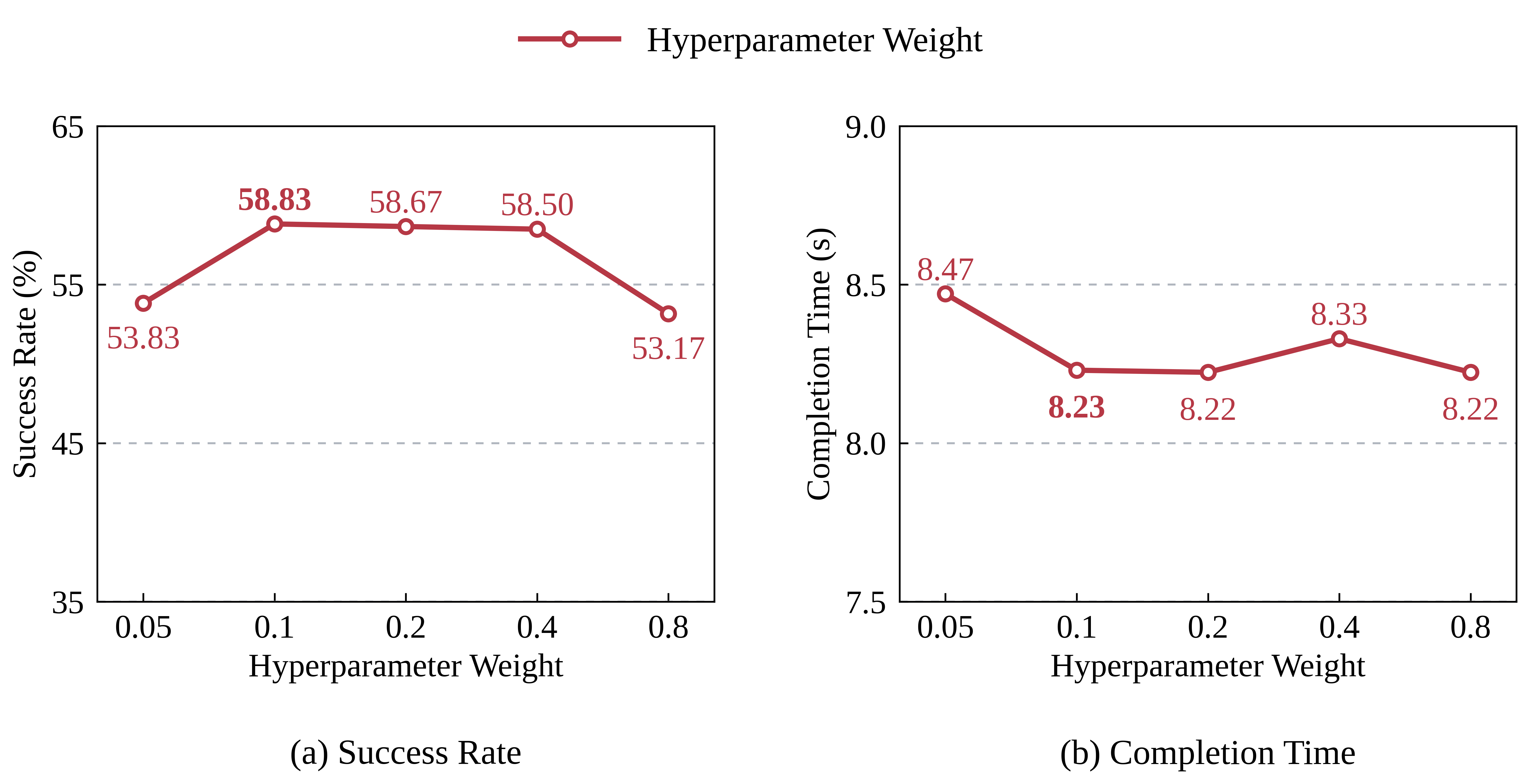}
  \caption{Effect of Hyper $\lambda_V$.}
  \label{fig:value_weight}
\end{figure}

\subsection{Effect of Value Loss Weight}  We evaluate the effect of the value loss weight $\lambda_V$. As shown in Fig.~\ref{fig:value_weight}, PhysVAM achieves the highest success rate of $58.83\%$ at $\lambda_V=0.1$, with an average execution time of $8.23$~s. Performance remains stable for $\lambda_V\in\{0.1,0.2,0.4\}$, while smaller or larger weights reduce the success rate. Finally, we chose $\lambda_V=0.1$ in all experiments to balance action generation and future-value learning.

\subsection{Detailed Ablation Results}
Tables~\ref{tab:detailed_value} and \ref{tab:detailed_motion} provide per-object results for the value-guided action generation and motion history ablations reported in Tables~\ref{tab:ablation_value} and \ref{tab:ablation_motion}, respectively.

\begin{table}[t]
    \centering
    \caption{
        \textbf{Detailed results of value-guided action generation ablations.}
        Per-object success rates across different value branch
        designs, conditioning representations, and prediction
        objectives. Best results are shown in bold.
    }
    \label{tab:detailed_value}

    \begingroup
    \large
    \setlength{\tabcolsep}{1pt}
    \renewcommand{\arraystretch}{1.25}

    \resizebox{\textwidth}{!}{
        \begin{tabular}{@{}l*{12}{c}@{}}
            \toprule

            \multirow{2}{*}{Methods}
            & \multicolumn{5}{c}{Moderate-speed}
            & \multicolumn{5}{c}{High-speed}
            & \multicolumn{2}{c}{Average} \\

            \cmidrule(lr){2-6}
            \cmidrule(lr){7-11}
            \cmidrule(l){12-13}

            & Avocado
            & Beer
            & Potato
            & Apple
            & Lemon
            & Can
            & D-Bottle
            & W-Bottle
            & Egg
            & Cup
            & SR (\%) $\uparrow$
            & Time (s) $\downarrow$ \\

            \midrule
            \multicolumn{13}{@{}l}{
                \textit{(a) Value branch}
            } \\

            Without Value Branch
            & \srval{48.33}{6.24}
            & \srval{36.67}{6.24}
            & \srval{56.67}{2.36}
            & \srval{41.67}{2.36}
            & \srval{75.00}{4.08}
            & \textbf{\srval{40.00}{12.25}}
            & \srval{85.00}{7.07}
            & \srval{71.67}{9.43}
            & \srval{16.67}{4.71}
            & \srval{33.33}{2.36}
            & \srval{50.50}{1.08}
            & \srval{9.65}{0.02} \\

            \midrule
            \multicolumn{13}{@{}l}{
                \textit{(b) Value representations}
            } \\

            Scalar Conditioning
            & \srval{53.33}{20.14}
            & \srval{30.00}{14.72}
            & \srval{45.00}{8.16}
            & \srval{36.67}{8.50}
            & \srval{46.67}{6.24}
            & \srval{8.33}{6.24}
            & \srval{73.33}{2.36}
            & \srval{71.67}{9.43}
            & \srval{21.67}{4.71}
            & \srval{21.67}{10.27}
            & \srval{40.83}{1.43}
            & \srval{9.02}{0.16} \\

            Binary Conditioning
            & \textbf{\srval{68.33}{2.36}}
            & \srval{25.00}{7.07}
            & \srval{46.67}{10.27}
            & \srval{30.00}{7.07}
            & \srval{53.33}{6.24}
            & \textbf{\srval{40.00}{4.08}}
            & \srval{83.33}{9.43}
            & \srval{80.00}{4.08}
            & \srval{13.33}{2.36}
            & \textbf{\srval{45.00}{14.72}}
            & \srval{48.50}{2.55}
            & \srval{8.85}{0.08} \\

            \midrule
            \multicolumn{13}{@{}l}{
                \textit{(c) Multi-horizon value prediction}
            } \\

            Single-Value (Single Query)
            & \srval{61.67}{11.79}
            & \srval{18.33}{6.24}
            & \srval{56.67}{14.34}
            & \srval{31.67}{4.70}
            & \srval{11.67}{6.27}
            & \srval{3.33}{4.71}
            & \srval{76.67}{2.36}
            & \srval{58.33}{4.71}
            & \srval{20.00}{10.80}
            & \srval{40.00}{0.00}
            & \srval{37.83}{3.30}
            & \srval{9.30}{0.10} \\

            Single-Value (Multiple Queries)
            & \srval{56.67}{6.24}
            & \srval{46.67}{6.24}
            & \srval{50.00}{4.08}
            & \srval{23.33}{4.71}
            & \srval{81.67}{15.46}
            & \srval{30.00}{10.80}
            & \srval{71.67}{8.50}
            & \textbf{\srval{96.67}{4.71}}
            & \srval{21.67}{6.24}
            & \srval{26.67}{4.71}
            & \srval{50.50}{0.71}
            & \srval{8.40}{0.13} \\

            \midrule
            \multicolumn{13}{@{}l}{
                \textit{(d) Future progress prediction}
            } \\

            Current-Value Prediction
            & \srval{65.00}{7.07}
            & \textbf{\srval{55.00}{8.16}}
            & \srval{41.67}{16.50}
            & \srval{40.00}{4.08}
            & \srval{38.33}{4.71}
            & \srval{26.67}{6.24}
            & \srval{75.00}{10.80}
            & \srval{81.67}{10.27}
            & \srval{8.33}{2.36}
            & \srval{33.33}{2.36}
            & \srval{46.50}{2.27}
            & \srval{9.02}{0.16} \\

            \midrule

            \textbf{PhysVAM (Full)}
            & \srval{60.00}{10.80}
            & \srval{41.67}{10.27}
            & \textbf{\srval{65.00}{7.07}}
            & \textbf{\srval{45.00}{4.08}}
            & \textbf{\srval{88.33}{6.24}}
            & \srval{36.67}{10.27}
            & \textbf{\srval{86.67}{8.50}}
            & \srval{93.33}{4.71}
            & \textbf{\srval{28.33}{8.50}}
            & \srval{43.33}{4.71}
            & \textbf{\srval{58.83}{1.65}}
            & \textbf{\srval{8.23}{0.13}} \\

            \bottomrule
        \end{tabular}
    }
    \endgroup
\end{table}

\noindent\textbf{Detailed Value Ablation Analysis.} As shown in Table~\ref{tab:detailed_value}, different value designs exhibit considerable
performance variations across objects. Although some variants achieve higher success rates on individual objects, PhysVAM obtains the highest average success rate and shortest execution time. Notably, multi-horizon future-value learning provides stronger overall performance than single-value prediction and current-value assessment, highlighting the benefit of explicitly modeling future task progress for dynamic manipulation.

\begin{table}[t!]
    \centering
    \caption{
        \textbf{Detailed results of object and robot motion history ablations.}
        Per-object success rates across different motion representations.
        All historical inputs span four frames.
        Best results are shown in bold.
    }
    \label{tab:detailed_motion}

    \begingroup
    \large
    \setlength{\tabcolsep}{1pt}
    \renewcommand{\arraystretch}{1.25}

    \resizebox{\textwidth}{!}{
        \begin{tabular}{@{}l*{12}{c}@{}}
            \toprule

            \multirow{2}{*}{Methods}
            & \multicolumn{5}{c}{Moderate-speed}
            & \multicolumn{5}{c}{High-speed}
            & \multicolumn{2}{c}{Average} \\

            \cmidrule(lr){2-6}
            \cmidrule(lr){7-11}
            \cmidrule(l){12-13}

            & Avocado
            & Beer
            & Potato
            & Apple
            & Lemon
            & Can
            & D-Bottle
            & W-Bottle
            & Egg
            & Cup
            & SR (\%) $\uparrow$
            & Time (s) $\downarrow$ \\

            \midrule
            \multicolumn{13}{@{}l}{
                \textit{(a) Object motion representation}
            } \\

            Without Object Motion History
            & \srval{0.00}{0.00}
            & \srval{18.33}{4.71}
            & \srval{1.67}{2.36}
            & \srval{43.33}{11.79}
            & \srval{0.00}{0.00}
            & \srval{0.00}{0.00}
            & \srval{78.33}{10.27}
            & \srval{6.67}{4.71}
            & \srval{10.00}{0.00}
            & \srval{13.33}{4.71}
            & \srval{17.17}{2.39}
            & \srval{9.65}{0.02} \\

            RGB History
            & \srval{5.00}{0.00}
            & \srval{23.33}{10.27}
            & \srval{15.00}{4.08}
            & \srval{18.33}{6.24}
            & \srval{6.67}{4.71}
            & \srval{0.00}{0.00}
            & \textbf{\srval{91.67}{4.71}}
            & \srval{18.33}{12.47}
            & \srval{21.67}{8.50}
            & \srval{23.33}{6.24}
            & \srval{22.33}{1.55}
            & \srval{9.27}{0.26} \\

            Object Position History
            & \srval{30.00}{4.08}
            & \srval{23.33}{4.71}
            & \srval{15.00}{8.16}
            & \srval{35.00}{4.08}
            & \srval{33.33}{6.24}
            & \srval{1.67}{2.36}
            & \srval{85.00}{4.08}
            & \srval{16.67}{6.24}
            & \srval{13.33}{8.50}
            & \textbf{\srval{53.33}{12.47}}
            & \srval{30.67}{2.01}
            & \srval{9.53}{0.03} \\

            Object Velocity History (Oracle)
            & \srval{6.67}{2.36}
            & \srval{21.67}{2.36}
            & \srval{41.67}{14.34}
            & \srval{36.67}{2.36}
            & \srval{26.67}{8.50}
            & \srval{3.33}{2.36}
            & \srval{88.33}{2.36}
            & \srval{76.67}{6.24}
            & \srval{23.33}{4.71}
            & \srval{15.00}{4.08}
            & \srval{34.00}{3.27}
            & \srval{9.00}{0.10} \\

            Object Velocity History (Estimated)
            & \srval{11.67}{6.24}
            & \textbf{\srval{41.67}{14.34}}
            & \srval{30.00}{12.25}
            & \srval{28.33}{4.71}
            & \srval{5.00}{4.08}
            & \srval{10.00}{7.07}
            & \srval{78.33}{4.71}
            & \srval{83.33}{4.71}
            & \srval{16.67}{9.43}
            & \srval{31.67}{9.43}
            & \srval{33.67}{0.85}
            & \srval{8.93}{0.04} \\

            \midrule
            \multicolumn{13}{@{}l}{
                \textit{(b) Robot motion history}
            } \\

            Without Robot Motion History
            & \srval{51.67}{11.79}
            & \srval{40.00}{10.80}
            & \srval{35.00}{12.25}
            & \srval{38.33}{4.71}
            & \srval{41.67}{2.36}
            & \srval{6.67}{2.36}
            & \srval{88.33}{2.36}
            & \srval{88.33}{8.50}
            & \srval{28.33}{4.71}
            & \srval{33.33}{11.79}
            & \srval{45.17}{2.05}
            & \srval{8.86}{0.04} \\

            \midrule

            \textbf{PhysVAM (Full)}
            & \textbf{\srval{60.00}{10.80}}
            & \textbf{\srval{41.67}{10.27}}
            & \textbf{\srval{65.00}{7.07}}
            & \textbf{\srval{45.00}{4.08}}
            & \textbf{\srval{88.33}{6.24}}
            & \textbf{\srval{36.67}{10.27}}
            & \srval{86.67}{8.50}
            & \textbf{\srval{93.33}{4.71}}
            & \textbf{\srval{28.33}{8.50}}
            & \srval{43.33}{4.71}
            & \textbf{\srval{58.83}{1.65}}
            & \textbf{\srval{8.23}{0.13}} \\

            \bottomrule
        \end{tabular}
    }
    \endgroup
\end{table}

\noindent\textbf{Detailed Motion History Analysis.}
As shown in Table~\ref{tab:detailed_motion}, motion representations exhibit varying effectiveness across object categories. Removing robot motion history
reduces the average success rate from $58.83\%$ to $45.17\%$, with particularly pronounced degradation on the potato, lemon, and can. Meanwhile, object
position and velocity histories provide complementary information, with their combination achieving the highest average success rate. Notably, PhysVAM achieves the best or tied-best results on eight of the ten object scenes, demonstrating the benefit of jointly modeling object and robot motion for dynamic manipulation.

\subsection{Detailed Real World Results}
\noindent\textbf{Per-object Real-world Results.} Table~\ref{tab:real_world_per_object} reports detailed success rates across five object instances. PhysVAM consistently outperforms both baselines, achieving gains of $15.0$--$40.0$ percentage points over PUMA across all stances. Its overall success rate reaches $64.17\%$, showing consistent effectiveness across different objects and dynamic scenarios.

\begin{table}[t!]
    \centering
    \caption{
        \textbf{Detailed per-object real-world manipulation results.}
        Success rates (SR) are reported with successful
        trials and total trials in parentheses. Overall Avg. is computed by equally weighting the three dynamic manipulation scenarios. Best results are shown in bold. 
    }
    \label{tab:real_world_per_object}

    \begingroup
    \normalsize
    \setlength{\tabcolsep}{2pt}
    \renewcommand{\arraystretch}{1.2}

    \resizebox{0.60\columnwidth}{!}{
    \begin{tabular}{ccccc}
        \toprule
        Scenario & Object
        & DynamicVLA & PUMA & PhysVAM \\
        \midrule

        \multirow{3}{*}{Conveyor}
        & Orange & 30.0 (6/20) & 35.0 (7/20)
        & \textbf{70.0 (14/20)} \\

        & Cube & 50.0 (10/20) & 60.0 (12/20)
        & \textbf{75.0 (15/20)} \\

        & Avg. & 40.0 & 47.5 & \textbf{72.5} \\

        \midrule

        \multirow{3}{*}{Turntable}
        & Bread & 25.0 (5/20) & 35.0 (7/20)
        & \textbf{60.0 (12/20)} \\

        & Cup & 20.0 (4/20) & 30.0 (6/20)
        & \textbf{70.0 (14/20)} \\

        & Avg. & 22.5 & 32.5 & \textbf{65.0} \\

        \midrule

        Rolling
        & Bottle & 25.0 (5/20) & 30.0 (6/20)
        & \textbf{55.0 (11/20)} \\

        \midrule

        \multicolumn{2}{c}{Overall Avg.}
        & 29.17 & 36.67 & \textbf{64.17} \\

        \bottomrule
    \end{tabular}
    }
    \endgroup
\end{table}

\end{document}